\documentclass[11pt]{article}

\usepackage[a4paper,margin=1in]{geometry}
\usepackage{iftex}
\usepackage{booktabs}
\usepackage{multirow}
\usepackage{graphicx}
\usepackage{xcolor}
\usepackage{array}
\usepackage{authblk}
\usepackage[numbers]{natbib}
\usepackage{xurl}
\usepackage[hidelinks]{hyperref}

\ifPDFTeX
  \usepackage[utf8]{inputenc}
  \usepackage[T1]{fontenc}
\fi

\ifXeTeX
  \usepackage{fontspec}
  \usepackage{ucharclasses}
  \newfontfamily\khmerfont[
    Path=fonts/,
    UprightFont = NotoSerifKhmer-Regular.ttf,
    BoldFont = NotoSerifKhmer-Bold.ttf,
    Script=Khmer
  ]{Noto Serif Khmer}
  \setTransitionsFor{Khmer}{\begingroup\khmerfont}{\endgroup}
\fi

\title{A Comparative Study of Language Models for Khmer Retrieval-Augmented Question Answering}

\author[1]{Sereiwathna Ros}
\author[2]{Phannet Pov}
\author[2]{Ratanaktepi Chhor}
\author[3]{Kimleang Ly}
\author[4,5]{Wan-Sup Cho}
\author[4,5]{Saksonita Khoeurn\thanks{Corresponding author: saksonita@chungbuk.ac.kr}}

\affil[1]{Department of Computer Science, Chungbuk National University, Cheongju, South Korea}
\affil[2]{Department of Big Data, Chungbuk National University, Cheongju, South Korea}
\affil[3]{General Department of Information and Communication Technology, Ministry of Post and Telecommunications, Phnom Penh, Cambodia}
\affil[4]{Department of Management Information Systems, Chungbuk National University, Cheongju, South Korea}
\affil[5]{BigDatalabs Co., Ltd, Cheongju, South Korea}

\date{}

\newcommand{\keywords}[1]{\par\noindent\textbf{Keywords: }#1}
\newcommand{\Description}[1]{}
\begin{document}
\maketitle

\begin{abstract}
Retrieval-Augmented Generation (RAG) has emerged as a promising paradigm for grounding large language model (LLM) outputs in retrieved evidence, thereby reducing hallucination and improving factual accuracy. Its efficacy, however, remains largely unexamined for low-resource, non-Latin-script languages such as Khmer. In this paper, we present a RAG-based question answering system for Khmer-language telecom-domain documents. We conduct a two-phase comparative evaluation. First, we benchmark three embedding models—BGE-M3 (567M), Jina-Embeddings-v3 (570M), and Qwen3-Embedding (597M)—for dense retrieval over Khmer documents. BGE-M3 consistently performs best, achieving a Hit Rate@3 of 0.285, File Hit Rate@3 of 0.700, MRR@3 of 0.221, and Precision@3 of 0.112, substantially outperforming the other retrievers. Second, using BGE-M3 as the selected retriever, we evaluate five generator backends—Qwen3 (8B), Qwen3.5 (9B), Sailor2-8B-Chat, SeaLLMs-v3-7B-Chat, and Llama-SEA-LION-v2-8B-IT—on a curated golden dataset of 200 Khmer question-answer pairs. To quantify system performance, we apply six RAGAS-inspired metrics: faithfulness, answer relevance, context relevance, factual correctness, answer similarity, and answer correctness. The results show no single model dominates across all metrics: Qwen3.5-9B achieves the highest faithfulness (0.859) and context relevance (0.726), Qwen3-8B attains the highest factual correctness (0.380), and SeaLLMs-v3-7B-Chat performs best on answer relevance (0.867), answer similarity (0.836), and answer correctness (0.599). These findings highlight that retriever choice remains a major bottleneck for Khmer RAG, while generator strengths vary depending on whether the priority is grounding, factual precision, or semantic similarity.
\end{abstract}

\keywords{retrieval-augmented generation, RAG evaluation, RAGAS metrics, Khmer question answering, Khmer NLP, local LLMs, dense retrieval, low-resource languages}

%% ====================================================================
\section{Introduction}
\label{sec:introduction}

Retrieval-Augmented Generation (RAG)~\cite{lewis2020rag} has become a common approach for question answering over domain-specific document collections because it combines external retrieval with the generative capabilities of Large Language Models (LLMs). In such systems, performance depends not only on the generator, but also on the quality of retrieval and on whether the generated answer remains grounded in the retrieved evidence. As a result, evaluating RAG is inherently multi-dimensional: a system may fail because relevant evidence is not retrieved, because the model does not use the retrieved context effectively, or because it hallucinates unsupported content. Existing evaluation practices, however, remain heavily shaped by English-centric settings and do not always transfer cleanly to low-resource languages or institutionally sensitive domains~\cite{gao2024ragsurvey,es2024ragas,roychowdhury2024ragmetrics}.

These limitations are especially important for Khmer document Question Answering (QA). Khmer is a low-resource language written in a complex Abugida script, with limited annotated resources and weak standardization for word segmentation. These characteristics introduce challenges at multiple stages of a RAG pipeline, including text extraction, document preprocessing, retrieval, and answer evaluation. In institutional settings, the problem is further compounded by the need for trustworthy responses grounded in authoritative documents, since hallucinated or weakly supported answers can undermine public confidence~\cite{ji2023hallucination}. Despite growing interest in multilingual and low-resource Natural Language Processing (NLP), it remains unclear which retrieval models, which locally deployable generators, and which automated evaluation signals are most suitable for Khmer-language RAG systems.

In this paper, we present a systematic study of retrieval-augmented question answering over Khmer institutional documents. Our focus is not only on end-to-end answer quality, but also on the interaction between retrieval, generation, and automated evaluation in a low-resource, non-Latin-script setting. We study a privacy-preserving RAG pipeline built for locally hosted deployment over Khmer telecom-domain documents, allowing us to examine the practical requirements of document-grounded question answering under data-sovereignty constraints.

Our contributions are threefold:
\begin{enumerate}
    \item We benchmark dense retrievers for Khmer document retrieval and analyze their effectiveness on noisy, domain-diverse institutional text.
    \item We compare five locally deployable generator models, including both general-purpose multilingual LLMs and Southeast Asian-focused models, to assess whether regional specialization yields measurable gains for Khmer question answering.
    \item We examine six adapted RAGAS-style metrics in this setting and discuss their usefulness and limitations for evaluating Khmer RAG pipelines.
\end{enumerate}

To support this study, we construct a gold evaluation set of 200 Khmer question--answer pairs derived from authoritative documents across multiple institutional subdomains. Using this benchmark, we provide an empirical account of retrieval quality, answer quality, and metric behavior in Khmer document QA. Our findings show that retriever choice has a substantial effect on downstream performance, while the relative strengths of generator models vary across grounding-oriented and similarity-oriented evaluation measures. More broadly, this paper highlights the need for language-aware RAG evaluation practices in low-resource settings, and offers evidence that methods validated in English should not be assumed to behave identically for Khmer.

\section{Related Work}
\label{sec:relatedwork}

RAG combines a retriever with an LLM so that answers are generated from retrieved evidence rather than from parametric memory alone~\cite{lewis2020rag}. In a typical pipeline, documents are collected, segmented into passages, indexed, and retrieved at inference time to condition answer generation~\cite{gao2024ragsurvey}. Because end-to-end performance depends jointly on retrieval quality and response grounding, RAG has become a widely used framework for knowledge-intensive tasks. Early systems often relied on dense retrieval methods such as Dense Passage Retrieval (DPR)~\cite{karpukhin2020dense}, while more recent multilingual retrievers such as BGE-M3~\cite{chen2024bge} aim to improve transfer across languages and scripts. These developments are especially relevant in low-resource settings, where retrieval can be degraded by limited training data, Optical Character Recognition (OCR) artifacts, and script-sensitive preprocessing challenges~\cite{hosseinbeigi2025advancing}.

Traditional reference-based metrics such as Bilingual Evaluation Understudy (BLEU) and Recall-Oriented Understudy for Gisting Evaluation (ROUGE) measure lexical overlap with gold references~\cite{papineni2002bleu,lin2004rouge}, while embedding-based metrics such as Bidirectional Encoder Representations from Transformers Score (BERTScore) better capture semantic similarity through contextualized representations~\cite{zhang2019bertscore}. However, these approaches are not fully adequate for RAG because they do not directly assess whether a response is supported by the retrieved evidence. This limitation is particularly important in low-resource settings, where high-quality reference sets are costly to construct and correct answers may show substantial lexical and syntactic variation.

Recent work therefore moves beyond answer similarity toward explicit assessment of grounding and context use. GEval demonstrates that LLM-as-a-judge methods can support flexible rubric-based evaluation~\cite{liu2023g}. RAGAS adapts this idea to RAG pipelines through metrics such as faithfulness, answer relevance, and context-related measures~\cite{es2024ragas,ragas_github}. Related benchmarks such as RGB further stress-test RAG systems under noisy retrieval and counterfactual conditions~\cite{chen2024benchmarking}. In an applied telecom setting, Roychowdhury et al.~\cite{roychowdhury2024ragmetrics} likewise report that grounding-oriented measures such as faithfulness and factual correctness align more closely with expert judgment than similarity-based metrics. Taken together, this literature suggests that RAG evaluation should account not only for output quality, but also for evidential support.

Although RAG and LLM evaluation have advanced rapidly, most evidence still comes from English and other high-resource languages. For Khmer, this creates an important gap: document processing and retrieval must contend with a non-Latin script, inconsistent word segmentation, OCR noise, and limited task-specific resources. Regionally focused models such as SEA-LION~\cite{ng2025sealionsoutheastasianlanguages} and Sailor2~\cite{dou2025sailor2sailingsoutheastasia} indicate growing support for Southeast Asian languages, but they do not by themselves establish how well retrieval models, generator models, and automated evaluation metrics behave in Khmer institutional QA settings.
% TODO (Phannet): Add 2--4 Khmer-specific citations here, e.g., prior work on Khmer NLP,

Ly et al. ~\cite{ly2024fine} conducted a study in which they prepared a dataset of questions and corresponding Khmer answers to perform fine-tuning experiments on large language models (LLMs) for the Khmer language. To evaluate the generated answers, the authors employed similarity-based metrics that compare the model outputs with reference answers. Specifically, ROUGE-1, ROUGE-2, and ROUGE-L were used to measure unigram overlap, bigram overlap, and the longest common subsequence, respectively. 
\begin{figure}[htbp]
\centering
\includegraphics[width=0.98\columnwidth]{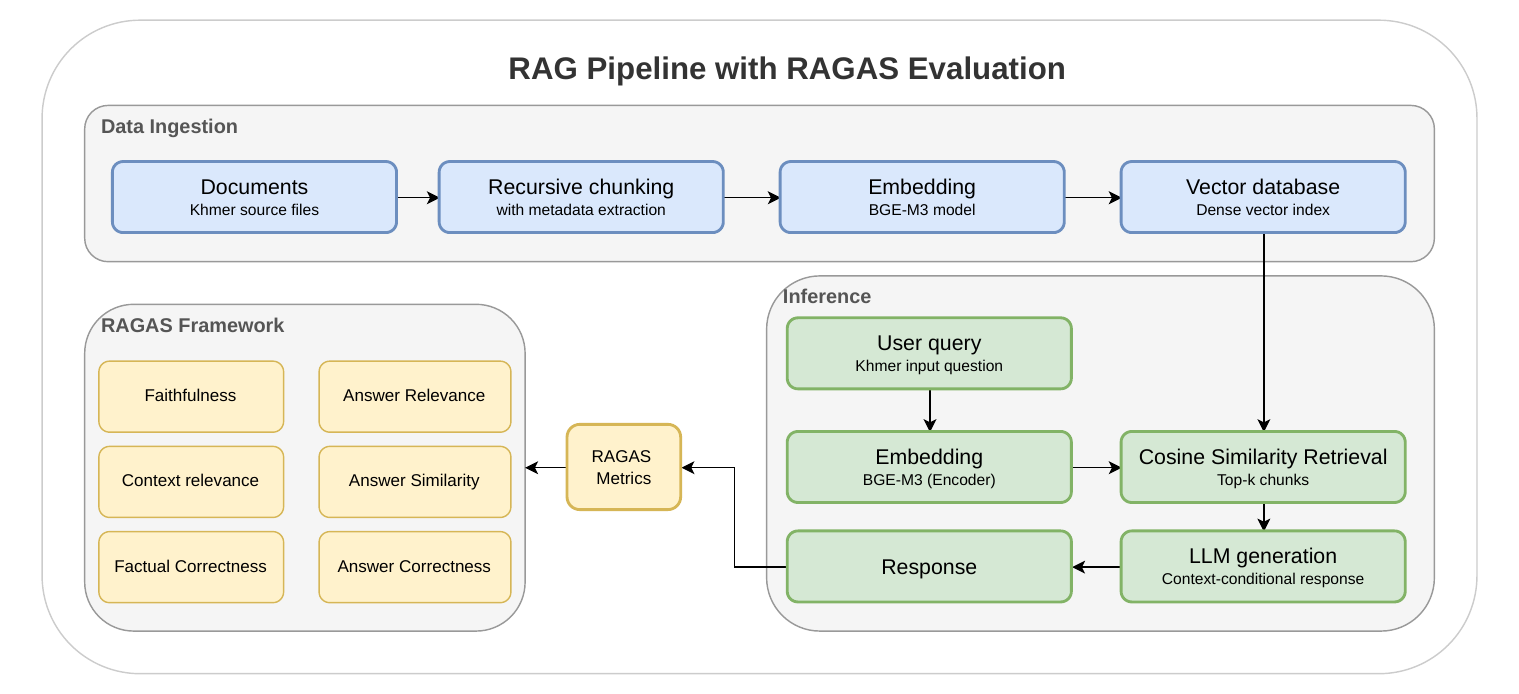}
\caption{System architecture of the RAG pipeline.}
\Description{RAG system architecture showing offline preprocessing and indexing, followed by online query embedding, retrieval, generation, and source-grounded response output.}
\label{fig:architecture}
\end{figure}
These metrics provide insights into how closely the generated responses align with the ground truth answers. In contrast, our work generates answers based on retrieved contextual information rather than relying on direct comparison with predefined target answers. Therefore, traditional similarity-based evaluation metrics may not fully capture the quality of the responses, as they do not adequately reflect the relevance and faithfulness of the generated answers to the provided context.

More broadly, prior work on Khmer and related low-resource language processing suggests that language-specific preprocessing and resource constraints can materially affect downstream system performance. Our study builds on this perspective by evaluating retrieval quality, answer generation, and automated RAG assessment jointly in a Khmer-language institutional document environment.

Overall, this work is positioned at the intersection of multilingual RAG, grounding-aware evaluation, and low-resource language processing. Unlike prior work centered on English, our focus is not only whether RAG works, but whether its retrieval and evaluation assumptions remain reliable in a Khmer-language setting.

%% ====================================================================
\section{Methodology}
\label{sec:methodology}
%% ====================================================================

Our experimental setup follows a standard RAG pipeline comprising a retriever followed by a generator module. Figure~\ref{fig:architecture} shows the schematic of our pipeline. The input to the system is a user query in Khmer or English, which is processed through dense retrieval followed by LLM-based answer generation conditioned on the retrieved context.

\subsection{Dataset}
\label{sec:dataset}

We collected open-source data from websites that publish official documents related to Information and Communication Technology (ICT). The data sources include notifications, guidelines, laws, announcements, decrees, sub-decrees, government documents, press releases, Q\&A documents, decisions, and general information. The corpus is mostly Khmer language documents with embedded English technical terms. The focus on closed-domain institutional documents enables this research to create a specific and verifiable knowledge domain, which is critical in building question answering systems in a low-resource language setting.

The documents are preprocessed into Markdown format and recursively chunked into segments that preserve semantic coherence. Each chunk carries metadata including the source document identifier and a unique chunk ID for provenance tracking. The resulting corpus contains over 7,000 chunks.

For evaluation, we curate a golden dataset comprising 200 question--answer pairs derived from the telecom domain document corpus. The questions span multiple domains and reflect realistic citizen queries written in Khmer. Each entry consists of: (1)~a question, (2)~a target answer (ground truth) composed by domain experts, and (3)~domain metadata (document ID, question ID, domain category) for stratified analysis.

\subsection{Retriever Models}
\label{sec:retriever}

The retriever module is based on dense passage retrieval. An encoder-based language model computes embeddings for both the query and the document chunks. For every query embedding, the retriever outputs the top-$k$ most similar chunk embeddings using cosine similarity.

We evaluate three embedding models to determine which provides the most effective dense retrieval for Khmer documents:
\begin{enumerate}
  \item \textbf{BGE-M3} (567M)~\cite{chen2024bge}: Supports multi-lingual, multi-functionality, and multi-granularity embeddings through self-knowledge distillation, achieving competitive performance across 100+ languages including Southeast Asian scripts. Served through Ollama~\cite{ollama2024}.
  \item \textbf{Jina-Embeddings-v3} (570M)~\cite{sturua2024jina}: An embedding model supporting 89+ languages with task-specific adapters for retrieval, classification, and semantic similarity. Served through Ollama.
  \item \textbf{Qwen3-Embedding} (597M)~\cite{qwen3embedding}: A compact embedding model from the Qwen family designed for semantic search across multiple languages. Served through Ollama.
\end{enumerate}

Document chunks are embedded in batches during the offline stage, and the complete vector database is serialized for efficient runtime loading. At query time, the top-$k$ chunks (default $k{=}3$) are selected and concatenated with similarity scores and source metadata into a structured context string. The retriever is evaluated using standard information retrieval metrics: Hit Rate@$k$, File-level Hit Rate@$k$, Mean Reciprocal Rank (MRR@$k$), and Precision@$k$.

\subsection{Generator Models}
\label{sec:generator}

Once the relevant context has been retrieved for a query, the query and context are passed to the LLM for generating the response. A system prompt constrains the LLM to use only the provided context and not generate information beyond it.

\begin{table}[htbp]
\caption{Summary of models used in the experimental setup.}
\label{tab:models}
\centering
\small
\begin{tabular}{@{}llll@{}}
\toprule
\textbf{Role} & \textbf{Model} & \textbf{Params} & \textbf{Runtime} \\
\midrule
Embedding & BGE-M3 & 567M & Ollama \\
Embedding & Jina-Embeddings-v3 & 570M & Ollama \\
Embedding & Qwen3-Embedding & 600M & Ollama \\
Generator & Qwen3 & 8B & Ollama \\
Generator & Qwen3.5 & 9B & Ollama \\
Generator & Sailor2-Chat & 8B & HuggingFace \\
Generator & SeaLLMs-v3-7B & 7B & HuggingFace \\
Generator & Llama-SEA-LION-v2 & 8B & HuggingFace \\
Judge & GPT-4o-mini & -- & OpenAI API \\
% Similarity & all-MiniLM-L6-v2 & 22M & Sentence-BERT \\
\bottomrule
\end{tabular}
\end{table}

We evaluate five LLM backends with different sizes and linguistic focus:
\begin{enumerate}
  \item \textbf{Qwen3 (8B)}~\cite{qwen3technicalreport}: A general-purpose model from the Qwen family with broad language coverage. Served locally via Ollama.
  \item \textbf{Qwen3.5 (9B)}~\cite{qwen35blog}: A newer-generation model from the Qwen family with improved instruction-following capabilities. Served locally via Ollama.
  \item \textbf{Sailor2-8B-Chat}~\cite{dou2025sailor2sailingsoutheastasia}: An 8B-parameter model specifically trained for Southeast Asian language understanding. Inference is performed using HuggingFace Transformers.
  \item \textbf{SeaLLMs-v3-7B-Chat}~\cite{damonlp2024seallm3}: A 7B chat model optimized for Southeast Asian languages. Inference is performed using HuggingFace Transformers.
  \item \textbf{Llama-SEA-LION-v2-8B-IT}~\cite{ng2025sealionsoutheastasianlanguages}: An 8B instruction-tuned SEA-LION family model targeting Southeast Asian multilingual use. Inference is performed using HuggingFace Transformers.
\end{enumerate}

All local inference is performed on a single machine with one NVIDIA GPU H200.

\subsection{Evaluation Metrics}
\label{sec:evaluation}

We focus on six adapted metrics inspired by the RAGAS framework~\cite{es2024ragas}. Higher value is better for all of them. The evaluation employs GPT-4o-mini~\cite{openai2024gpt4o} as the LLM judge and BGE-M3~\cite{chen2024bge} for computing semantic similarity. We use the notation of Roychowdhury et al.~\cite{roychowdhury2024ragmetrics}: given question $q$ and context $c(q)$ retrieved from the corpus, the LLM generates answer $a(q)$. The ground truth answer is denoted $gt(q)$.

\begin{itemize}
  \item \textbf{Faithfulness ($\mathit{FaiFul}$):} Checks if the generated statements from $a(q)$ are present in the retrieved context $c(q)$ through verdicts; the ratio of valid verdicts $V$ to total number of statements $S(q)$ is the answer's faithfulness:
  \begin{equation}
  \mathit{FaiFul} = \frac{|V|}{|S(q)|}
  \end{equation}

  \item \textbf{Answer Relevance ($\mathit{AnsRel}$):} The average cosine similarity of the user's question $q$ with $N{=}3$ generated questions $\tilde{q}_i$, using $a(q)$ as reference, is the answer relevance:
  \begin{equation}
  \mathit{AnsRel} = \frac{1}{N} \sum_{i=1}^{N} \mathrm{sim}(E(q), E(\tilde{q}_i))
  \end{equation}

  \item \textbf{Context Relevance ($\mathit{ConRel}$):} The semantic similarity between the question $q$ and the retrieved context $c(q)$ computed using BGE-M3 embeddings:
  \begin{equation}
  \mathit{ConRel} = \mathrm{sim}(E(q), E(c(q)))
  \end{equation}

  \item \textbf{Answer Similarity ($\mathit{AnsSim}$):} The similarity between the embedding of $a(q)$ and the embedding of $gt(q)$:
  \begin{equation}
  \mathit{AnsSim} = \mathrm{sim}(E(a(q)), E(gt(q)))
  \end{equation}

\begin{table}[htbp]
\caption{Retriever evaluation results on the 200-question golden dataset. Bold values indicate the best score per metric. $\uparrow$ indicates higher is better.}
\label{tab:retriever_results}
\centering
\small
\begin{tabular}{@{}lccc@{}}
\toprule
\textbf{Metric} ($\uparrow$) & \textbf{BGE-M3} & \textbf{Jina v3} & \textbf{Qwen3-Emb} \\
                              & \textbf{(567M)} & \textbf{(570M)}                 & \textbf{(597M)} \\
\midrule
Hit Rate@1       & \textbf{0.170} & 0.075 & 0.110 \\
Hit Rate@3       & \textbf{0.285} & 0.135 & 0.175 \\
Hit Rate@5       & \textbf{0.355} & 0.170 & 0.205 \\
Hit Rate@10      & \textbf{0.385} & 0.205 & 0.225 \\
\midrule
File Hit Rate@3  & \textbf{0.700} & 0.485 & 0.525 \\
File Hit Rate@5  & \textbf{0.805} & 0.575 & 0.620 \\
File Hit Rate@10 & \textbf{0.835} & 0.645 & 0.675 \\
\midrule
MRR@3            & \textbf{0.221} & 0.099 & 0.141 \\
Precision@3      & \textbf{0.112} & 0.055 & 0.065 \\
Cosine Sim (top-1) & 0.704 & \textbf{0.759} & 0.686 \\
\bottomrule
\end{tabular}
\end{table}

  \item \textbf{Factual Correctness ($\mathit{FacCor}$):} The F1-Score of statements in $a(q)$ classified as True Positive (TP), False Positive (FP) and False Negative (FN) by the LLM judge with respect to $gt(q)$:
  \begin{equation}
  \mathit{FacCor} = \frac{|\mathrm{TP}|}{|\mathrm{TP}| + 0.5 \times (|\mathrm{FP}| + |\mathrm{FN}|)}
  \end{equation}

  \item \textbf{Answer Correctness ($\mathit{AnsCor}$):} Determines correctness of $a(q)$ with respect to $gt(q)$ as a weighted sum of factual correctness and answer similarity:
  \begin{equation}
  \mathit{AnsCor} = w_1 \times \mathit{FacCor} + w_2 \times \mathit{AnsSim}
  \end{equation}
  with weights $[w_1, w_2] = [0.5, 0.5]$ in our implementation.
\end{itemize}

For each of the 200 questions, the evaluation pipeline: (1)~retrieves the top-3 document chunks from the vector database, (2)~generates an LLM answer conditioned on the retrieved context, (3)~records the question, target answer, LLM answer, and retrieved chunks, and (4)~computes the six metrics. The evaluation is conducted separately for each of the five LLM backends. For Qwen3 and Qwen3.5, evaluation is run via Ollama. For Sailor2-8B-Chat, SeaLLMs-v3-7B-Chat, and Llama-SEA-LION-v2-8B-IT, inference is performed using HuggingFace Transformers. The generation evaluation uses BGE-M3 as the retriever, which was identified as the best-performing embedding model in our retriever comparison (Section~\ref{sec:retriever_results}).

%% ====================================================================
\section{Results and Discussion}
\label{sec:results}
%% ====================================================================

The results of the RAG evaluation are presented in two parts. First, we compare the three embedding models for retrieval quality (Table~\ref{tab:retriever_results}). Then, using BGE-M3 as the selected retriever, we evaluate the five generator models on the six RAGAS metrics (Table~\ref{tab:main_results}).

\subsection{Retriever Evaluation}
\label{sec:retriever_results}

Table~\ref{tab:retriever_results} presents the retrieval performance of the three embedding models evaluated on the 200-question golden dataset. We report Hit Rate (the fraction of queries for which the correct chunk appears in the top-$k$ results), File-level Hit Rate (the fraction for which a chunk from the correct source file appears), Mean Reciprocal Rank (MRR), and Precision at different values of $k$.

\begin{table}[htbp]
\caption{RAGAS metrics for the 200-question golden dataset with $k{=}3$ retrieved contexts (BGE-M3 retriever). Numbers are mean scores. Bold values indicate the best score per metric. $\uparrow$ indicates higher is better.}
\label{tab:main_results}
\centering
\small
\resizebox{\columnwidth}{!}{%
\begin{tabular}{@{}lccccc@{}}
\toprule
\textbf{Metric} ($\uparrow$) & \textbf{Qwen3 (8B)} & \textbf{Qwen3.5 (9B)} & \textbf{Sailor2 (8B)} & \textbf{SeaLLMs (7B)} & \textbf{Llama-SEA-LION (8B)} \\
\midrule
$\mathit{FaiFul}$ & 0.780 & \textbf{0.859} & 0.758 & 0.846 & 0.556 \\
$\mathit{AnsRel}$ & 0.808 & 0.779 & 0.802 & \textbf{0.867} & 0.831 \\
$\mathit{ConRel}$ & 0.718 & \textbf{0.726} & 0.717 & 0.718 & 0.718 \\
$\mathit{FacCor}$ & \textbf{0.380} & 0.303 & 0.258 & 0.352 & 0.217 \\
$\mathit{AnsSim}$ & 0.648 & 0.661 & 0.606 & \textbf{0.836} & 0.766 \\
$\mathit{AnsCor}$ & 0.521 & 0.480 & 0.432 & \textbf{0.599} & 0.488 \\
\bottomrule
\end{tabular}
}
\end{table}

BGE-M3 achieves the highest scores across all retrieval effectiveness metrics, with a Hit Rate@3 of 0.285---more than double that of Jina-Embeddings-v3 (0.135) and substantially higher than Qwen3-Embedding (0.175). The advantage is even more pronounced at the file level: BGE-M3 retrieves a chunk from the correct source file 70\% of the time at $k{=}3$, compared to 52.5\% for Qwen3-Embedding and 48.5\% for Jina v3. Interestingly, Jina v3 achieves the highest top-1 cosine similarity (0.759), yet this does not translate into better retrieval accuracy, consistent with cautions about interpreting raw cosine similarity as a quality signal~\cite{steck2024cosine}. Based on these results, we select BGE-M3 as the retriever for the generator evaluation.

\subsection{Generator Evaluation}
\label{sec:generator_results}

Table~\ref{tab:main_results} presents the RAGAS metrics for the five generator models on the 200-question golden dataset using BGE-M3 as the retriever.

\subsection{Discussion on Metrics}
\label{sec:metric_discussion}

We discuss our findings about the six RAGAS metrics and their behaviour in the context of Khmer  document QA.

\paragraph{Faithfulness ($\mathit{FaiFul}$).}
Qwen3.5-9B achieves the highest faithfulness score of 0.859, with SeaLLMs-v3-7B-Chat close behind at 0.846. Qwen3-8B and Sailor2-8B remain in a similar middle range (0.780 and 0.758), whereas Llama-SEA-LION-v2-8B-IT drops markedly to 0.556. This ranking suggests that strong multilingual or regional coverage alone is insufficient; what matters is whether the model can stay tightly grounded in the retrieved evidence. The result is consistent with the findings of Roychowdhury et al.~\cite{roychowdhury2024ragmetrics}, who report that faithfulness is generally concordant with manual evaluation by subject matter experts.

\begin{table}[htbp]
\caption{Summary of key findings per metric.}
\label{tab:key_findings}
\centering
\small
\begin{tabular}{@{}l>{\raggedright\arraybackslash}p{12cm}@{}}
\toprule
\textbf{Metric} & \textbf{Key Conclusion} \\
\midrule
$\mathit{FaiFul}$ & \textbf{Qwen3.5 wins}: most grounded in context (0.859). \\
$\mathit{AnsRel}$ & \textbf{SeaLLMs wins}: strongest question-level relevance (0.867). \\
$\mathit{ConRel}$ & \textbf{Near tie}: retrieval quality dominates; LLM impact is small. \\
$\mathit{FacCor}$ & \textbf{Qwen3 wins}: highest factual accuracy (0.380). \\
$\mathit{AnsSim}$ & \textbf{SeaLLMs wins}: closest to ground truth semantically (0.836). \\
$\mathit{AnsCor}$ & \textbf{SeaLLMs wins}: best overall balance of accuracy and semantics (0.599). \\
\bottomrule
\end{tabular}
\end{table}

\paragraph{Answer Relevance ($\mathit{AnsRel}$).}
SeaLLMs-v3-7B-Chat achieves the highest answer relevance (0.867), followed by Llama-SEA-LION-v2-8B-IT (0.831), Qwen3-8B (0.808), Sailor2-8B (0.802), and Qwen3.5-9B (0.779). Unlike the earlier three-model comparison, the inclusion of two additional Southeast Asian-focused models shows that regional specialization can improve topical responsiveness to Khmer questions. However, as noted in the literature~\cite{steck2024cosine, roychowdhury2024ragmetrics}, this metric relies on cosine similarity between generated and original question embeddings and should therefore be interpreted cautiously as a relative signal rather than an absolute measure.

\paragraph{Context Relevance ($\mathit{ConRel}$).}
All five models achieve nearly identical context relevance scores, ranging only from 0.717 to 0.726, with Qwen3.5-9B slightly ahead at 0.726. This is expected because context relevance primarily reflects retrieval quality, which is determined by the shared BGE-M3 embedding model and is largely independent of the LLM backend. The tight clustering confirms that retrieval quality remains a system-level bottleneck rather than a differentiating property of the generators themselves. As observed by Roychowdhury et al.~\cite{roychowdhury2024ragmetrics}, context relevance is mainly indicative and dependent on context length, making it difficult to interpret as a standalone measure of answer quality.

\paragraph{Factual Correctness ($\mathit{FacCor}$).}
Qwen3-8B achieves the highest factual correctness (0.380), with SeaLLMs-v3-7B-Chat second at 0.352, followed by Qwen3.5-9B (0.303), Sailor2-8B (0.258), and Llama-SEA-LION-v2-8B-IT (0.217). The overall low scores across all models reflect the difficulty of the task: documents contain highly specific information such as phone numbers, URLs, and legal article numbers, where partial matches yield low F1 scores. Importantly, the ranking differs from faithfulness: Qwen3.5-9B is best grounded in retrieved context, but Qwen3-8B more accurately reproduces factual details from the ground truth. Together with faithfulness, this metric has been found to be most aligned with expert evaluation~\cite{roychowdhury2024ragmetrics} and remains one of the most informative measures in our setting.

\paragraph{Answer Similarity ($\mathit{AnsSim}$) and Answer Correctness ($\mathit{AnsCor}$).}
SeaLLMs-v3-7B-Chat achieves the highest answer similarity (0.836) and also the highest composite answer correctness (0.599), indicating the strongest overall balance between semantic closeness to the reference answer and factual adequacy. Llama-SEA-LION-v2-8B-IT also scores relatively high on answer similarity (0.766) but falls back to 0.488 on answer correctness because of its weak factual correctness. By contrast, Qwen3-8B remains competitive on answer correctness (0.521) despite lower answer similarity, because its stronger factual correctness compensates for the gap. This again shows that semantic similarity and factual accuracy are distinct dimensions of quality. Table~\ref{tab:key_findings} summarizes the key conclusions.

In summary, our results indicate that of these metrics, $\mathit{FaiFul}$ and $\mathit{FacCor}$ (and hence $\mathit{AnsCor}$) are perhaps best aligned with human expert judgment for our domain; scores for $\mathit{AnsSim}$, $\mathit{AnsRel}$ and $\mathit{ConRel}$ are subject to inherent variations as discussed above. This echoes the findings of Roychowdhury et al.~\cite{roychowdhury2024ragmetrics} for the telecom domain.

\subsection{Retriever Model Selection}
\label{sec:retriever_discussion}

A key finding is that BGE-M3 substantially outperforms both Jina-Embeddings-v3 and Qwen3-Embedding across all retrieval effectiveness metrics. At $k{=}3$, BGE-M3 achieves more than double the Hit Rate of Jina v3 (0.285 vs.\ 0.135) and a 63\% higher Hit Rate than Qwen3-Embedding (0.285 vs.\ 0.175). The gap widens at the file level, where BGE-M3 retrieves from the correct source file 70\% of the time compared to 52.5\% (Qwen3-Embedding) and 48.5\% (Jina v3).

An interesting observation is that Jina v3 achieves the highest mean cosine similarity for its top-1 retrieved chunk (0.759), yet produces the worst retrieval accuracy. This highlights a known limitation of relying on raw cosine similarity as a retrieval quality indicator~\cite{steck2024cosine}: a model may assign high similarity scores to semantically related but non-matching passages. BGE-M3's self-knowledge distillation approach appears to produce embeddings that better discriminate between truly relevant and merely related content for Khmer documents.

Nevertheless, even the best retriever (BGE-M3) achieves a relatively modest Hit Rate@3 of 0.285, indicating that retrieval quality remains a primary bottleneck in the overall RAG pipeline. This is reflected in the tightly clustered $\mathit{ConRel}$ scores ($\sim$0.72) observed across all five generator models.

\subsection{Language Focus vs.\ General Multilingual Models}
\label{sec:model_analysis}

With the addition of SeaLLMs-v3-7B-Chat and Llama-SEA-LION-v2-8B-IT, the comparison between general-purpose and Southeast Asian language-focused models becomes more nuanced. Regional specialization does not uniformly help or hurt Khmer RAG performance. SeaLLMs is the strongest model on answer relevance, answer similarity, and answer correctness, while Qwen3.5-9B remains best on faithfulness and Qwen3-8B remains best on factual correctness.

Within the Qwen family, an interesting trade-off remains: Qwen3.5-9B achieves higher faithfulness (0.859 vs.\ 0.780) and answer similarity (0.661 vs.\ 0.648), while Qwen3-8B leads on factual correctness (0.380 vs.\ 0.303) and answer correctness (0.521 vs.\ 0.480). This suggests that the newer, slightly larger Qwen3.5 model is better at grounding answers in retrieved context, while Qwen3 more precisely reproduces factual details from the ground truth.

The three Southeast Asian-focused models also separate clearly from one another. SeaLLMs appears to transfer well to Khmer answer generation, whereas Sailor2 and especially Llama-SEA-LION struggle more on grounding-oriented metrics. Several hypotheses may explain this spread: (1)~the amount and quality of Khmer data may differ substantially across SEA-focused pretraining corpora; (2)~instruction-following ability under constrained RAG prompting may matter more than regional coverage alone; and (3)~differences in tokenizer behaviour, chat templates, and HuggingFace inference settings may influence answer style and grounding. The broader conclusion is that regional language targeting can help, but the effect is strongly model-specific rather than guaranteed across all models in the same family.

From a deployment perspective, model choice depends on the target objective. For applications where faithfulness to retrieved context is paramount, Qwen3.5-9B is the best choice. For applications where factual accuracy matters most, Qwen3-8B is preferable. For applications prioritizing overall answer quality and semantic closeness to the target response, SeaLLMs-v3-7B-Chat is the strongest option among the evaluated models.

A key limitation of this study is that all reported scores are based on automated evaluation. Although prior work in the telecom domain suggests that faithfulness and factual correctness align more closely with expert judgment than similarity-based metrics, we did not conduct a dedicated human evaluation for Khmer in this study. Therefore, our conclusions about metric reliability should be interpreted as evidence from automated analysis rather than definitive validation against human annotations.

% \subsection{Challenges of Evaluation in Khmer}
% \label{sec:challenges}

% Several factors complicate the use of RAGAS metrics for Khmer:
% \begin{itemize}
%   \item \textbf{Script complexity.} Khmer uses an Abugida writing system where vowels attach to consonants in complex ways, complicating string matching and tokenization required by metrics like $\mathit{FaiFul}$ and $\mathit{FacCor}$.
%   \item \textbf{Lack of word boundaries.} Unlike English, Khmer does not use spaces to delimit words, complicating sentence segmentation for metrics like $\mathit{ConRel}$.
%   \item \textbf{LLM judge limitations.} The judge model (GPT-4o-mini) was primarily trained on high-resource languages and may not accurately assess Khmer text quality, potentially underestimating all models.
%   % \item \textbf{Embedding mismatch.} The Sentence-BERT model (all-MiniLM-L6-v2) used for $\mathit{AnsSim}$ and $\mathit{AnsRel}$ was trained primarily on English data, which may attenuate similarity scores for Khmer text pairs.
% \end{itemize}

% These challenges mirror concerns raised for domain-specific evaluation~\cite{roychowdhury2024ragmetrics} and suggest that verdict-based metrics ($\mathit{FaiFul}$, $\mathit{FacCor}$) are more reliable for Khmer than similarity-based metrics ($\mathit{AnsRel}$, $\mathit{AnsSim}$).

\subsection{Limitations}

\begin{itemize}
  \item \textbf{Fixed retrieval.} The current system uses a single dense retrieval pass with fixed $k{=}3$. Adaptive retrieval or iterative refinement could improve context quality.
  \item \textbf{Automated evaluation bias.} All metrics rely on LLM judges and English-centric embedding models, potentially introducing systematic bias for Khmer evaluation.
  \item \textbf{Limited ground truth.} The 200-question golden dataset, while carefully curated, may not capture the full distribution of citizen queries.
  \item \textbf{No human evaluation.} We did not conduct human evaluation of answer quality, which would provide a more reliable assessment.
  \item \textbf{Inference setup differences.} Sailor2-8B was evaluated with a different inference framework (HuggingFace) than the Qwen models (Ollama), which may introduce confounding factors.
\end{itemize}

%% ====================================================================
\section{Conclusion and Future Work}
\label{sec:conclusion}
%% ====================================================================

In this work, we presented a retrieval-augmented generation system for Khmer document question answering using locally deployed language models. We conducted a two-fold comparative evaluation: a retriever comparison of three embedding models, and a generator comparison of five LLMs (Qwen3-8B, Qwen3.5-9B, Sailor2-8B, SeaLLMs-v3-7B-Chat, and Llama-SEA-LION-v2-8B-IT) on 200 question--answer pairs across six RAGAS-inspired metrics. Our results show that verdict-based RAGAS metrics ($\mathit{FaiFul}$ and $\mathit{FacCor}$) provide more reliable evaluation signals for Khmer text than similarity-based metrics ($\mathit{AnsRel}$ and $\mathit{AnsSim}$), which remain sensitive to embedding-model choice and the inherent limitations of cosine similarity. This is consistent with findings from the telecom domain~\cite{roychowdhury2024ragmetrics} and suggests that these concerns also extend to low-resource, non-Latin-script language settings. Among the retrievers, BGE-M3 substantially outperforms Jina-Embeddings-v3 and Qwen3-Embedding on all retrieval metrics, achieving more than double the Hit Rate@3 of Jina v3. Notably, high cosine similarity alone, as observed with Jina v3, does not necessarily translate into strong retrieval accuracy, reinforcing the importance of task-specific retriever evaluation. For generation, performance cannot be explained by a simple general-purpose-versus-regional distinction. Qwen3.5-9B achieves the highest faithfulness (0.859), Qwen3-8B leads in factual correctness (0.380), and SeaLLMs-v3-7B-Chat performs best on answer relevance (0.867), answer similarity (0.836), and answer correctness (0.599), whereas Llama-SEA-LION-v2-8B-IT underperforms on faithfulness and factual correctness. These findings indicate that regional specialization can benefit Khmer question answering, but its impact is strongly model-dependent. Additionally, retrieval quality emerges as a significant bottleneck, as reflected in the modest Hit Rate@3 (0.285) and the tightly clustered context relevance scores ($\sim$0.72) across all five generator models.

Future work will focus on: (1)~improving retrieval through hybrid dense+sparse methods and query expansion; (2)~incorporating Khmer-specific tokenization and embedding models to improve both retrieval and evaluation quality; (3)~conducting human evaluation studies with citizens to validate automated metrics against expert judgments; (4)~exploring domain adaptation through instruction fine-tuning of smaller models on Khmer  data, which has been shown to improve metric concordance with expert evaluation~\cite{roychowdhury2024ragmetrics}; and (5)~extending the system to support cross-lingual QA where questions and documents may be in different languages.

\bibliographystyle{unsrtnat}
\bibliography{references}

@inproceedings{ly2024fine,
  title={Fine-Tuning for Question Answering in Low-Resource Languages: A Case Study on Khmer},
  author={Ly, Kimleang and Valy, Dona and Kong, Phutphalla},
  booktitle={2024 17th International Congress on Advanced Applied Informatics (IIAI-AAI-Winter)},
  pages={162--165},
  year={2024},
  publisher={IEEE},
  address={Kitakyushu, Japan}
}

@misc{qwen3technicalreport,
      title={Qwen3 Technical Report}, 
      author={Qwen Team},
      year={2025},
      eprint={2505.09388},
      archivePrefix={arXiv},
      primaryClass={cs.CL},
      url={https://arxiv.org/abs/2505.09388}, 
}

@misc{ng2025sealionsoutheastasianlanguages,
      title={SEA-LION: Southeast Asian Languages in One Network}, 
      author={Raymond Ng and Thanh Ngan Nguyen and Yuli Huang and Ngee Chia Tai and Wai Yi Leong and Wei Qi Leong and Xianbin Yong and Jian Gang Ngui and Yosephine Susanto and Nicholas Cheng and Hamsawardhini Rengarajan and Peerat Limkonchotiwat and Adithya Venkatadri Hulagadri and Kok Wai Teng and Yeo Yeow Tong and Bryan Siow and Wei Yi Teo and Wayne Lau and Choon Meng Tan and Brandon Ong and Zhi Hao Ong and Jann Railey Montalan and Adwin Chan and Sajeban Antonyrex and Ren Lee and Esther Choa and David Ong Tat-Wee and Bing Jie Darius Liu and William Chandra Tjhi and Erik Cambria and Leslie Teo},
      year={2025},
      eprint={2504.05747},
      archivePrefix={arXiv},
      primaryClass={cs.CL},
      url={https://arxiv.org/abs/2504.05747}, 
}

@misc{dou2025sailor2sailingsoutheastasia,
      title={Sailor2: Sailing in South-East Asia with Inclusive Multilingual LLMs}, 
      author={Longxu Dou and Qian Liu and Fan Zhou and Changyu Chen and Zili Wang and Ziqi Jin and Zichen Liu and Tongyao Zhu and Cunxiao Du and Penghui Yang and Haonan Wang and Jiaheng Liu and Yongchi Zhao and Xiachong Feng and Xin Mao and Man Tsung Yeung and Kunat Pipatanakul and Fajri Koto and Min Si Thu and Hynek Kydlíček and Zeyi Liu and Qunshu Lin and Sittipong Sripaisarnmongkol and Kridtaphad Sae-Khow and Nirattisai Thongchim and Taechawat Konkaew and Narong Borijindargoon and Anh Dao and Matichon Maneegard and Phakphum Artkaew and Zheng-Xin Yong and Quan Nguyen and Wannaphong Phatthiyaphaibun and Hoang H. Tran and Mike Zhang and Shiqi Chen and Tianyu Pang and Chao Du and Xinyi Wan and Wei Lu and Min Lin},
      year={2025},
      eprint={2502.12982},
      archivePrefix={arXiv},
      primaryClass={cs.CL},
      url={https://arxiv.org/abs/2502.12982}, 
}

@article{qwen3embedding,
  title={Qwen3 Embedding: Advancing Text Embedding and Reranking Through Foundation Models},
  author={Zhang, Yanzhao and Li, Mingxin and Long, Dingkun and Zhang, Xin and Lin, Huan and Yang, Baosong and Xie, Pengjun and Yang, An and Liu, Dayiheng and Lin, Junyang and Huang, Fei and Zhou, Jingren},
  journal={arXiv preprint arXiv:2506.05176},
  year={2025}
}

@misc{qwen35blog,
    title = {Qwen3.5: Accelerating Productivity with Native Multimodal Agents},
    url = {https://qwen.ai/blog?id=qwen3.5},
    author = {Qwen Team},
    month = {February},
    year = {2026}
}

@misc{chen2024bge,
  author        = {Jianlv Chen and Shitao Xiao and Peitian Zhang and Kun Luo and Defu Lian and Zheng Liu},
  title         = {{BGE} {M3}-Embedding: Multi-Lingual, Multi-Functionality, Multi-Granularity Text Embeddings Through Self-Knowledge Distillation},
  year          = {2024},
  eprint        = {2402.03216},
  archivePrefix = {arXiv},
  primaryClass  = {cs.CL},
}

@misc{hosseinbeigi2025advancing,
  title={Advancing Retrieval-Augmented Generation for Persian: Development of Language Models, Comprehensive Benchmarks, and Best Practices for Optimization},
  author={Hosseinbeigi, Sara Bourbour and Asghari, Sina and Kashani, Mohammad Ali Seif and Shalchian, Mohammad Hossein and Abbasi, Mohammad Amin},
  year={2025},
  eprint={2501.04858},
  archivePrefix={arXiv},
  primaryClass={cs.CL}
}

@inproceedings{liu2023g,
  title={G-eval: NLG evaluation using gpt-4 with better human alignment},
  author={Liu, Yang and Iter, Dan and Xu, Yichong and Wang, Shuohang and Xu, Ruochen and Zhu, Chenguang},
  booktitle={Proceedings of the 2023 Conference on Empirical Methods in Natural Language Processing},
  pages={2511--2522},
  year={2023},
  publisher={Association for Computational Linguistics},
  address={Singapore}
}

@misc{ragas_github,
  author       = {{Exploding Gradients}},
  title        = {RAGAS: Retrieval Augmented Generation Assessment},
  howpublished = {\url{https://github.com/vibrantlabsai/ragas}},
  note         = {GitHub repository, accessed 2026-03-12},
  year         = {2026},
}

@inproceedings{chen2024benchmarking,
  title={Benchmarking large language models in retrieval-augmented generation},
  author={Chen, Jiawei and Lin, Hongyu and Han, Xianpei and Sun, Le},
  booktitle={Proceedings of the AAAI Conference on Artificial Intelligence},
  volume={38},
  pages={17754--17762},
  publisher={AAAI Press},
  address={Vancouver, Canada},
  year={2024}
}

@inproceedings{es2024ragas,
  title={Ragas: Automated evaluation of retrieval augmented generation},
  author={Es, Shahul and James, Jithin and Anke, Luis Espinosa and Schockaert, Steven},
  booktitle={Proceedings of the 18th Conference of the European Chapter of the Association for Computational Linguistics: System Demonstrations},
  pages={150--158},
  publisher={Association for Computational Linguistics},
  address={St. Julian's, Malta},
  year={2024}
}

@inproceedings{lewis2020rag,
  author    = {Patrick Lewis and Ethan Perez and Aleksandra Piktus and Fabio Petroni and Vladimir Karpukhin and Naman Goyal and Heinrich K{"u}ttler and Mike Lewis and Wen-tau Yih and Tim Rockt{"a}schel and Sebastian Riedel and Douwe Kiela},
  title     = {Retrieval-Augmented Generation for Knowledge-Intensive {NLP} Tasks},
  booktitle = {Advances in Neural Information Processing Systems (NeurIPS)},
  year      = {2020},
  volume    = {33},
  pages     = {9459--9474},
  publisher = {Curran Associates, Inc.},
  address   = {Red Hook, NY, USA},
}

@misc{ollama2024,
  author       = {{Ollama}},
  title        = {Ollama: Run Large Language Models Locally},
  howpublished = {\url{https://ollama.com}},
  year         = {2024},
}

@inproceedings{karpukhin2020dense,
  author    = {Vladimir Karpukhin and Barlas Oguz and Sewon Min and Patrick Lewis and Ledell Wu and Sergey Edunov and Danqi Chen and Wen-tau Yih},
  title     = {Dense Passage Retrieval for Open-Domain Question Answering},
  booktitle = {Proceedings of the 2020 Conference on Empirical Methods in Natural Language Processing (EMNLP)},
  year      = {2020},
  pages     = {6769--6781},
  publisher = {Association for Computational Linguistics},
  address   = {Online},
}

@misc{gao2024ragsurvey,
  author        = {Yunfan Gao and Yun Xiong and Xinyu Gao and Kangxiang Jia and Jinliu Pan and Yuxi Bi and Yi Dai and Jiawei Sun and Meng Wang and Haofen Wang},
  title         = {Retrieval-Augmented Generation for Large Language Models: A Survey},
  year          = {2024},
  eprint        = {2312.10997},
  archivePrefix = {arXiv},
  primaryClass  = {cs.CL},
}

@misc{damonlp2024seallm3,
  author        = {Wenxuan Zhang and Hou Pong Chan and Yiran Zhao and Mahani Aljunied and Jianyu Wang and Chaoqun Liu and Yue Deng and Zhiqiang Hu and Weiwen Xu and Yew Ken Chia and Xin Li and Lidong Bing},
  title         = {SeaLLMs 3: Open Foundation and Chat Multilingual Large Language Models for Southeast Asian Languages},
  year          = {2024},
  eprint        = {2407.19672},
  archivePrefix = {arXiv},
  primaryClass  = {cs.CL},
}

@article{ji2023hallucination,
  author    = {Ziwei Ji and Nayeon Lee and Rita Frieske and Tiezheng Yu and Dan Su and Yan Xu and Etsuko Ishii and Yejin Bang and Andrea Madotto and Pascale Fung},
  title     = {Survey of Hallucination in Natural Language Generation},
  journal   = {ACM Computing Surveys},
  year      = {2023},
  volume    = {55},
  number    = {12},
  pages     = {1--38},
}

@misc{openai2024gpt4o,
  author       = {{OpenAI}},
  title        = {{GPT-4o} Mini},
  howpublished = {\url{https://openai.com/index/gpt-4o-mini-advancing-cost-efficient-intelligence}},
  year         = {2024},
}

@inproceedings{roychowdhury2024ragmetrics,
  author    = {Sujoy Roychowdhury and Sumit Soman and H G Ranjani and Neeraj Gunda and Vansh Chhabra and Sai Krishna Bala},
  title     = {Evaluation of {RAG} Metrics for Question Answering in the Telecom Domain},
  booktitle = {ICML 2024 Workshop on Foundation Models in the Wild},
  year      = {2024},
  pages     = {1--7},
  publisher = {PMLR},
  address   = {Vienna, Austria},
  note      = {arXiv:2407.12873},
}

@inproceedings{steck2024cosine,
  author    = {Harald Steck and Chaitanya Ekanadham and Nathan Kallus},
  title     = {Is Cosine-Similarity of Embeddings Really About Similarity?},
  booktitle = {Companion Proceedings of the ACM on Web Conference 2024},
  pages     = {887--890},
  year      = {2024},
  publisher = {Association for Computing Machinery},
  address   = {New York, NY, USA},
}

@inproceedings{papineni2002bleu,
  author    = {Kishore Papineni and Salim Roukos and Todd Ward and Wei-Jing Zhu},
  title     = {{BLEU}: A Method for Automatic Evaluation of Machine Translation},
  booktitle = {Proceedings of the 40th Annual Meeting of the ACL},
  pages     = {311--318},
  year      = {2002},
  publisher = {Association for Computational Linguistics},
  address   = {Philadelphia, PA, USA},
}

@inproceedings{lin2004rouge,
  author    = {Chin-Yew Lin},
  title     = {{ROUGE}: A Package for Automatic Evaluation of Summaries},
  booktitle = {Text Summarization Branches Out},
  pages     = {74--81},
  year      = {2004},
  publisher = {Association for Computational Linguistics},
  address   = {Barcelona, Spain},
}

@misc{zhang2019bertscore,
  author        = {Tianyi Zhang and Varsha Kishore and Felix Wu and Kilian Q. Weinberger and Yoav Artzi},
  title         = {{BERTScore}: Evaluating Text Generation with {BERT}},
  year          = {2019},
  eprint        = {1904.09675},
  archivePrefix = {arXiv},
  primaryClass  = {cs.CL},
}

@misc{sturua2024jina,
  author        = {Saba Sturua and Isabelle Mohr and Mohammad Kalim Akram and Michael G{\"u}nther and Bo Wang and Markus Krimmel and Feng Wang and Georgios Mastrapas and Andreas Koukounas and Nan Wang and Han Xiao},
  title         = {jina-embeddings-v3: Multilingual Embeddings With Task LoRA},
  year          = {2024},
  eprint        = {2409.10173},
  archivePrefix = {arXiv},
  primaryClass  = {cs.CL},
}

\appendix

\section{Computation of RAGAS Metrics}
\label{sec:appendix_metrics}

We refer the reader to Es et al.~\cite{es2024ragas} and Roychowdhury et al.~\cite{roychowdhury2024ragmetrics} for details on the metrics, but for completeness, we summarize the prompts and steps involved in our adapted implementation. The notation is as follows: given question $q$ and context $c(q)$ retrieved from the corpus, the LLM generates answer $a(q)$. The ground truth answer is $gt(q)$.

\subsection{Faithfulness (FaiFul)}

Faithfulness is computed in two steps. First, the LLM judge decomposes $a(q)$ into atomic statements $S(q)$ using the prompt: \emph{``Given a question and answer, create one or more statements from each sentence in the given answer.''}  Second, for each statement $s \in S(q)$, the LLM judge determines a binary verdict $v(s, c(q))$ indicating whether the statement is supported by the context. Faithfulness is the ratio of supported verdicts to total statements (Equation~1).

\subsection{Answer Relevance (AnsRel)}

The LLM judge generates $N$ questions from $a(q)$ using the prompt: \emph{``Generate a question for the given answer.''} The cosine similarity between the embedding of the original question $q$ and each generated question $\tilde{q}_i$ is computed, and the average is reported as answer relevance (Equation~2).

\subsection{Context Relevance (ConRel)}

In our implementation, context relevance is computed directly as the cosine similarity between the BGE-M3 embedding of the question $q$ and the BGE-M3 embedding of the concatenated retrieved context $c(q)$. Unlike the original LLM-based extraction variant of RAGAS, this implementation does not require the judge model to extract relevant sentences from the context.

\subsection{Factual Correctness (FacCor) and Answer Correctness (AnsCor)}

The LLM judge classifies statements from $a(q)$ and $gt(q)$ into True Positives (TP), False Positives (FP), and False Negatives (FN). Factual correctness is the F1 score (Equation~5). Answer correctness is the weighted sum of factual correctness and answer similarity (Equation~6).

\subsection{Answer Similarity (AnsSim)}

Answer similarity is the cosine similarity between BGE M3-embedding of $a(q)$ and $gt(q)$ (Equation~4). This metric does not involve the LLM judge.
\begingroup
\tiny
\setlength{\tabcolsep}{1.5pt}
\renewcommand{\arraystretch}{0.78}

\par\medskip
\refstepcounter{table}
\noindent\textbf{Table~\thetable. Single-sample metric trace in English.}
\label{tab:single_sample_metric_trace_english}
\par\smallskip
\centering
\begin{tabular}{|>{\raggedright\arraybackslash}p{0.22\linewidth}|>{\raggedright\arraybackslash}p{0.73\linewidth}|}
\hline
\textbf{Component} & \textbf{Description} \\
\hline

\textbf{Summary metrics} &
For this example, the final scores are $\mathrm{CtxRel}{=}0.726295$, $\mathrm{AnsSim}{=}0.800418$, $\mathrm{Faith}{=}1.000$, $\mathrm{AnsRel}{=}0.797417$, $\mathrm{FacCor}{=}0.500$, and $\mathrm{AnsCor}{=}0.650209$. \\
\hline

\textbf{Question and reference answer} &
\textbf{Q:} Which code cancels unwanted mobile supplementary services? \quad
\textbf{GT:} Dial *1200\# to unsubscribe from unwanted add-on services. \\
\hline

\textbf{Generated answer} &
The code to disconnect VAS without balance check, subscriber ID check, top-up, or customer-service contact is \textbf{*1200\#}. \par
Dial *1200\# and press send to drop add-ons without balance or ID steps, as stated in the notice. \\
\hline

\textbf{Faithfulness} &
\textbf{Generated statements (4):} \par
1. The code used to unsubscribe from mobile supplementary services (VAS) is \textbf{code *1200\#}. \par
2. Code *1200\# can be used without requiring balance top-up actions or identity verification. \par
3. To unsubscribe from supplementary services, the user dials *1200\# and presses send. \par
4. Unsubscribing from supplementary services does not require checking balance or contacting customer service. \par
\textbf{Verification verdicts:} Statement 1: Yes; Statement 2: Yes; Statement 3: Yes; Statement 4: Yes. \par
\textbf{Computation:} supported statements $= 4$, total statements $= 4$, therefore $\mathit{FaiFul} = 4/4 = 1.000$. \\
\hline

\textbf{Answer relevance} &
Three reverse questions were generated from the answer. Their cosine similarities to the original question were $0.845994$, $0.735499$, and $0.810756$, giving a mean answer-relevance score of $0.797417$. \\
\hline

\textbf{Context relevance} &
The cosine similarity between the question embedding and the retrieved-context embedding was $0.726295$. \\
\hline

\textbf{Factual correctness} &
\textbf{True positives (TP):} \par
1. Code *1200\# can be used to unsubscribe from unwanted mobile supplementary services. \par
\textbf{False positives (FP):} \par
1. The code to unsubscribe from mobile supplementary services (VAS) without needing a balance check, subscriber identity verification, top-up, or customer-service contact is \textbf{code *1200\#}. \par
2. By dialing *1200\# and pressing send, users can unsubscribe from supplementary services without balance top-up or identity-verification actions, as announced in the public notice. \par
\textbf{False negatives (FN):} None. \par
\textbf{Computation:} TP $= 1$, FP $= 2$, FN $= 0$; precision $= 1/(1+2) = 1/3$, recall $= 1/(1+0) = 1$, and $F_1 = 2 \times (1/3) \times 1 / (1/3 + 1) = 0.500$. \\
\hline

\textbf{Answer similarity / correctness} &
\textbf{Answer similarity:} cosine similarity between the generated-answer embedding and the reference-answer embedding was $0.800418$. \par
\textbf{Answer correctness formula:} $\mathit{AnsCor} = 0.5 \times \mathit{FacCor} + 0.5 \times \mathit{AnsSim}$. \par
\textbf{Values used:} $\mathit{FacCor} = 0.500000$ and $\mathit{AnsSim} = 0.800418$. \par
\textbf{Computation:} $\mathit{AnsCor} = 0.5 \times 0.500000 + 0.5 \times 0.800418 = 0.650209$. \\
\hline

\end{tabular}
\par\medskip
\endgroup

\end{document}